\documentclass[]{article}

\usepackage[english]{babel}
\usepackage{xcolor}
\usepackage{natbib}
\usepackage{mathtools,amsthm,amsfonts,amssymb}%

\usepackage{geometry}
\usepackage{enumitem}%
\usepackage{tikz}
\usetikzlibrary{arrows.meta, positioning, shapes.geometric}
\newcommand{\Eion}[1][j]{\mathsf{E}_{#1}} 
\newcommand{\Gmax}[1][j]{\mathsf{G}_{#1}} 
\newcommand{\gfun}[1][j]{g_{#1}} 
\newcommand{\Istim}{I_{\text{stim}}} 
\newcommand{\ICf}[1][k]{\mathsf{IC50}_{#1}} 
\newcommand{\conc}{\mathsf{D}} 
\newcommand{\dinput}{\mathsf{X}} 
\newcommand{\inhibfun}[1][j]{k_{#1}} 
\newcommand{\obs}{y} 
\newcommand{\pinnweights}{\theta} 
\newcommand{\hyperweights}{\Gamma} 
\newcommand{\approxU}[1][]{v_{\text{#1}}} 
\newcommand{\Loss}[1][]{\mathcal{L}^{\text{#1}}} 
\newcommand{\rmd}{{\rm{d}}} 
\newcommand{\deriv}[1]{\frac{\rmd #1}{\rmd t}} 
\newcommand{\normsq}[1]{\left\| #1 \right\|^2} 
\newcommand{\lw}[1]{w^{\text{#1}}} 

\newcommand{\keywords}[1]{\bgroup\noindent\textbf{Keywords:}
#1\egroup}%

\title{HyperSBINN: A hypernetwork-Enhanced Systems Biology-Informed Neural Network  for Efficient Drug Cardiosafety Assessment} 

\author{\footnotesize Inass Soukarieh$^{2, 3}$,	Gerhard Hessler$^{1}$, Hervé Minoux$^{2}$,\\ 
\footnotesize Marcel Mohr$^{1}$, Friedemann Schmidt$^{1}$,  Jan Wenzel$^{1}$,\\ 
\footnotesize Pierre Barbillon$^{3}$, Hugo Gangloff$^{3}$, Pierre Gloaguen$^{4, \ast}$.
\\
{\tiny$^{1}$ Sanofi, R\&D Preclinical Safety, Industriepark Hoechst, Frankfurt am Main, Germany}\\
{\tiny$^{2}$ Sanofi, Digital R\&D, 13 Quai Jules Guesde, Vitry-Sur-Seine, France}\\
{\tiny$^{3}$ Université Paris-Saclay, AgroParisTech, INRAE, UMR MIA
Paris-Saclay, 75005, Paris, France}\\
{\tiny$^{4}$ Université Bretagne Sud, UMR CNRS 6205, LMBA, F-56000 Vannes, France}\\
{\tiny$^\ast$pierre.gloaguen@univ-ubs.fr;}}

\date{}
\begin{document}

\maketitle

\begin{abstract}
Mathematical modeling in systems toxicology enables a comprehensive understanding of the effects of pharmaceutical substances on cardiac health. However, the complexity of these models limits their widespread application in early drug discovery. In this paper, we introduce a novel approach to solving parameterized models of cardiac action potentials by combining meta-learning techniques with Systems Biology-Informed Neural Networks (SBINNs). The proposed method, hyperSBINN, effectively addresses the challenge of predicting the effects of various compounds at different concentrations on cardiac action potentials, outperforming traditional differential equation solvers in speed. Our model efficiently handles scenarios with limited data and complex parameterized differential equations. The hyperSBINN model demonstrates robust performance in predicting APD90 values, indicating its potential as a reliable tool for modeling cardiac electrophysiology and aiding in preclinical drug development. This framework represents an advancement in computational modeling, offering a scalable and efficient solution for simulating and understanding complex biological systems.\\
\keywords{System-Biology informed neural network, hypernetwork and Cardiac action potential}
\end{abstract}

\section{Introduction}

The heart is one of the most important target organs of human drug toxicity. 
Consequently, modern drug discovery strives to screen out potential hazards early amongst multiple drug candidates, before an optimized drug candidate is selected and enters clinical trials. 
Toxicity testing has historically relied on evaluation of the adverse effects of drugs in dedicated animal studies to predict human safety of drug exposure. With the implementation of novel in silico predictive approaches, animal-free and cost-effective approaches are increasingly being developed to augment and ultimately replace animal experimentation (\cite{jenkinson2020practical}). 
Amongst others, the cardiac action potential (CAP) (\cite{andras2021cardiac}) is considered as a key parameter to monitor closely for undesired drug effects. CAP is a periodic and transient change of the membrane potential of heart cells. 
It is instantiated by a group of specialized pacemaker cells, which have automatic action potential generation capabilities. 
The voltage change is created by transport of charged ions through ion channels within the membrane. 
In healthy and undisturbed condition, the action potential passes along the cell membrane causing heart cells to contract periodically. 
Upon exposure to certain drugs, the morphology and frequency of the cardiac action potential can be disturbed, making this parameter an important marker for drug toxicity to the heart.  CAP is tightly controlled by proper functioning of multiple ion channels expressed on cardiac myocytes. 
The ionic currents through channels maintain regular heart beat. Mathematical models of cardiac electrophysiology usually start from the modeling of ionic current dynamics in single cells, coupled with models for the propagation of an action potential in tissue, and ultimately also mechanical parameters, such as contractility, cardiac output or blood pressure. 
The general approach to cardiac electrophysiology applies a set of coupled differential equations (\cite{trayanova2024computational, mayourian2018introduction}). 
Drugs may have the potential to alter the ion fluxes through undesired blocks of one or even multiple channels. These "off-target" effects have been found to be quite common, particularly with the human Ether-à-go-go-Related Gene (hERG) potassium channel (\cite{lee2019computational}).  Effects of drugs on ion channels can be mathematically expressed by partial block of the channels, leading to a reduced or stopped ionic current, and eventually implying alterations of the cardiac action potential(\cite{gerlach2001channel, carmeliet1998antiarrhythmic, andras2021cardiac}).

CAP modeling often employs systems of ordinary (ODEs) or partial differential equations (PDEs) such as the Hodgkin-Huxley equations or the Fitzhugh-Nagumo equations \cite{corrias2011ionic, williams2015web}. 
However, these models can be computationally expensive, limiting their practical utility in the early drug development process where numerous drug candidates or even drug candidate libraries have to be assessed in parallel to guide drug development optimally. 
To address this challenge, we propose testing Physics-Informed Neural Networks (PINNs) as an innovative deep learning tool for solving CAP model (\cite{raissi2019physics, cuomo2022scientific, krishnapriyan2021characterizing})). 
PINNs combine the precision of governing physical laws with the efficiency of deep learning.

Classical PINNs solve differential equations by embedding physical laws into the neural network framework through a specialized loss function. This integration allows PINNs to leverage the strengths of machine learning, such as automatic differentiation and optimization, to solve complex scientific and engineering problems \cite{raissi2019physics}. 
A significant advancement in the application of PINNs is their use in handling stiff ODE systems, which are characterized by solutions that can change rapidly and require careful numerical treatment (\cite{ sbinn, ji2021stiff, o2022stochastic}. 
Traditional numerical methods for stiff ODEs, like implicit solvers, can be computationally expensive and challenging to implement. 
PINNs offer a promising alternative by directly incorporating the stiffness of the system into the training process, thus providing a more efficient and scalable solution.

Recent work has started to explore the use of PINNs to analyze cardiac electrophysiology models. For instance, \cite{sahli2020physics} employed PINNs to solve the isotropic diffusion equation using in silico data, estimating action potential arrival times and conduction velocity maps. \cite{grandits2021learning} advanced this work by solving the anisotropic eikonal equation to learn about fiber orientations and conductivity tensors, though their study was based on synthetic data and data from only one patient. 
In a closely related study, \cite{herrero2022ep} applied PINNs with the monodomain equation to estimate electrophysiology parameters from sparse transmembrane potential maps, utilizing the Aliev-Panfilov model as described by \cite{aliev1996simple}, which lacks biological mechanism. Most recently, \cite{chiu2024characterisation} proposed a novel PINNs framework to predict how anti-arrhythmic drugs modulate myocardial electrophysiology parameters using in vitro optical mapping data and a biophysically detailed Fenton-Karma model. 

Though algorithmic progress in machine learning is undisputed, the acquisition of suitable training data remains a challenge, leading to limited information available to inform models. 
Data-driven machine learning techniques, such as deep neural networks, tend to fail under these circumstances. Physics-Informed Machine Learning (PiML) addresses this by blending data-driven machine learning with physics-based simulations. 
This hybrid approach is transforming problem-solving in science, engineering, and medicine \cite{alber2019integrating}. At the core of PiML are PINNs, a novel "meshfree" approach that leverages deep learning to solve PDEs commonly used to describe various phenomena in science and engineering. 
PINNs incorporate physical domain knowledge as soft constraints on an empirical loss function, optimized using existing machine learning training methodologies.

However, standard PINNs typically solve for a single set of parametrizations, meaning they are configured to solve the differential equations with specific fixed parameters. 
Therefore, when working in domains that require evaluating solutions across multiple parameter values, it is essential to adjust the approach to accommodate varying parameters. 
Two primary strategies for handling varying parametrizations in PINNs are prevalent:

\begin{enumerate}
    \item Retraining the network: This involves training a new PINN for each distinct set of parametrizations, which is costly due to the time-intensive nature of training.
    \item Including parameters as inputs: This approach increases the complexity of the PINN by directly propagating parametrizations into the network, resulting in decreased operational speed.
\end{enumerate}

This study introduces a third approach, named hyperSBINN (for hyper System biology informed neural network). 
This method mimicks hyperPINN, developed by \cite{de2021hyperpinn}, which adopts a meta-learning approach, utilizing a separate network (hypernetwork) to understand how parameters influence the equation. 
This allows the primary network responsible for solving the equation to remain simple and efficient.

When PDEs or ODEs are parameterized by a function that specifies initial or boundary conditions (the input function), operator learning methods—such as deep operator networks or Fourier neural operators, which learn the mapping from the input function to the solution of the PDE or ODE—have been applied in systems biology with success \citep{li2024new,centofanti2024learning,pellegrini2025learning}. 
In contrast, our system differs in that we use scalar parameters to parameterize the ODEs.
Given that the system under study assumes consistent initial conditions and involves a scalar parametrization, we focus on the hyperSBINN approach, even though an operator learning method could also accommodate scalar parameters (see \cite{huang2024operator}, for instance.

Our method is  illustrated on a highly validated mathematical CAP model presented by \cite{mohr2022accurate}. 
The parametrization variables of the ODE systems pertain to the compound's half-maximal inhibitory concentration (IC50) for a selection of four ionic channels: hERG potassium channel, hKv4.3 potassium channel, hCav1.2 L-type calcium channel, and hNav1.5 sodium channel. 
Each set of these four parameters characterizes the inhibitory potency of an individual compound on these selected channels. 
Given our objective of testing multiple compounds simultaneously, we encounter challenges in parametrized physics-informed neural networks,{  such as solving complex differential equations with a high number of parameters, handling high-dimensional parameters space, and ensuring accurate predictions across various parameter combinations }.

Our approach highlights the significance and effectiveness of employing hypernetworks to train the parameter space of IC50 values. We compare our findings with other machine learning techniques, to demonstrate the efficacy of hyperSBINN.

\section{Cardiac action potential model}
\label{sec:CAP:model}

Our study focuses on the modelling of the cardiac action potential (CAP) under different drug configurations.
We focus on the model proposed by \cite{mohr2022accurate}, which we briefly recall here, and fully detail in Appendix \ref{Appendix:CAP}.

In our context, the effect of a drug on CAP is supposed to depend only on its concentration (denoted $\conc$) and on the inhibitory effect of the drug on four ionic channels. 
The inhibitory effect of the $k$-th ($1\leqslant k \leqslant 4$) ionic channel is denoted by $\ICf$.
We denote $\dinput = \lbrace \conc, \ICf[1], \dots, \ICf[4] \rbrace$ the five controlled characteristics of the drug.
The CAP $V(t, \dinput)$ at time $t$ under the configuration $\dinput$ is related to $J = 8$ ionic currents (listed in the Appendix \ref{Appendix:CAP}), denoted by $I_j(t, \dinput)$ ($1\leqslant j \leqslant 8$) through the ordinary differential equation (ODE) \footnote{This is the classical equation for
electrical properties of a single cardiac cell, where we set the cell capacitance to 1, as in \cite{mohr2022accurate}.}
\begin{equation}
\label{eq:CAP:ODE}
     \deriv{V}(t, \dinput) = -\left(\Istim(t) + \sum_{j = 1}^8 I_j(t, \dinput) \right)\,,
\end{equation}
where $\Istim(t)$ is a known externally applied stimulus current.
The generic formulation of an ionic current $I_j(t, \dinput)$ from a channel $j$ is based on Ohm’s law and can be stated as 
\begin{equation}
\label{eq:genereic:ionic:current}
    I_j(t, \dinput)= \Gmax \cdot \inhibfun(\conc, \ICf) \cdot \gfun\left(t, V(t, \dinput)\right)\cdot (V(t, \dinput)-\Eion)\,,
\end{equation}
where:
\begin{itemize}
    \item $\Eion$ and $\Gmax$ is the known reversal potential and the maximal conductance, respectively, for channel $j$;
    \item $\inhibfun(\conc, \ICf)$ is a function describing the inhibition factor for channel $j$. This function is either always 1 (no inhibition), or takes the following form for the four ionic channels listed in the Appendix:
    $$
    \inhibfun(\conc, \ICf) = \left(1 + \frac{\conc}{\ICf}\right)^{-1}\,.
    $$
    \item $\gfun(t, V(t, \dinput))$ is the fraction of conducting channels, which depends on voltage and a set of kinetic parameters. All functions $\gfun$ are defined thanks to 13 solutions of ordinary differential equations depending on $V(t, \dinput)$. The 13 solutions of these ODEs lies in the interval [0, 1] and are referred to as \textit{gate functions}.
\end{itemize}
The complexity of the proposed models lies in the expressions of $V(t, \dinput)$ and the different $\gfun(t, V(t, \dinput))$, which requires solving a (nonlinear) system of 14 ODEs (given by Equations \eqref{eq:CAP:ODE} and \eqref{ODE:f-NaF}-\eqref{ODE:h-KS}, all listed in the Appendix \ref{Appendix:CAP}), whose solution has a nonlinear dependence on $\dinput$.
We denote by $u(t, \dinput)$ the function taking values in $\mathbb{R}^{14}$ which is the solution of our system of ODEs.
The next section depicts the framework to obtain a fast approximation of this function for every $t$ and $\dinput$ without using an ODE solver for each configuration $\dinput$.

\section[Hyper-SBINN]{System Biology-Informed Neural networks}
\label{sec:hyperSBINN}

\subsection{Overview of Physics-Informed Neural Networks}
\label{sec:PINN}

In this first part, we focus on the approximation of the solution for a given $\dinput$ (thus this variable is omitted in the notations).
Consider the problem of approximating a function $u(t) \in \mathbb{R}^d$, for $t\in [0, T]$ (where $T$ is a fixed time horizon), such that $u(t)$ is the solution of a nonlinear system of differential equations, expressed as:
\begin{equation}
\label{eq:ODE:system}
\begin{array}{rl}
    u(0) &= \mathsf{u_0}\,\\
    \deriv{u}(t) &= f(u(t))\,,
\end{array}
\end{equation}
where $f(\cdot)$ is a known non linear function and $\mathsf{u_0}$ is the known initial condition.
We suppose moreover that we have a set of measurements $\obs_{1:n} := \obs_1, \dots, \obs_n$ from the target function\footnote{Still, in this section, observations would come from a single drug configuration $\dinput$.} at times $t_{1:n} := t_1, \dots, t_n$ .

Physics-informed neural networks (PINNs, \citealp{raissi2019physics}), approximates the solution $u$ by a function $\approxU(t, \pinnweights)$, which is a neural network parameterized by a set of weights and biases $\pinnweights$. 
Neural networks provide a flexible set of arbitrarily complex functions that performs well in approximating non linear functions and whose derivatives (with respect to $t$ or $\pinnweights$) can be easily computed by automatic differentiation in standard machine learning softwares.
For a given neural network architecture\footnote{Defined by the number of layers, the number of neurons per layers, the connections between layers, and the activation functions.}, the approximation problem boils down to learn $\pinnweights$ that minimized some loss function $\Loss(\pinnweights)$. 
In our context, a "good" approximation is a function $\approxU(t, \pinnweights)$ such that:
\begin{itemize}
    \item at observation times $t_1, \dots, t_n$, the observations are well approximated, \textit{i.e.}, for all $1\leqslant i \leqslant n$,  $\approxU(t_i, \pinnweights) \approx y_i$.
    \item the approximation satisfies the initial condition, \textit{i.e.} $\approxU(0, \pinnweights) \approx \mathsf{u_0}$.
    \item the approximation satisfies Equation \eqref{eq:ODE:system}, \textit{i.e.}, for all $t \in [0, T]$,  
    $$\deriv{\approxU}(t ,\pinnweights)  \approx f(\approxU(t ,\pinnweights))\,.$$ 
\end{itemize}
These three objectives result in the following loss for the PINN:
\begin{equation}
\label{eq:PINN:loss}
\Loss[PINN](\pinnweights) = \lw{Data}\Loss[\text{Data}](\pinnweights) + 
\lw{IC}\Loss[IC](\pinnweights) + 
\lw{ODE}\Loss[ODE](\pinnweights)\,,
\end{equation}
where $\lw{Data},  \lw{IC}$ and $ \lw{ODE}$ are weights chosen by the user and:
\begin{align*}
    \Loss[Data](\pinnweights) &= \frac{1}{n}\sum_{i=1}^n \normsq{\approxU(t_i,\pinnweights) - y_i}\,,& \text{Data loss}\\
    \Loss[IC](\pinnweights) &= \normsq{\approxU(0,\pinnweights) - \mathsf{u_0}}\,, &\text{Initial condition loss}\\
    \Loss[ODE](\pinnweights) &= \frac{1}{m}\sum_{j=1}^m \normsq{\deriv{\approxU}(\tau_j,\pinnweights) - f(\approxU(\tau_j, \pinnweights))}\,. &\text{ODE loss}
\end{align*}
The first two losses are  mean squared errors at the observed values of the function. 
The third loss, evaluates the adequacy of the derivative of the approximation to the ODE model, at times $\tau_1, \dots, \tau_m$ that are chosen by the user.
These times are the analogous of the discretization times in a scheme use by a traditional ODE solver.

By optimizing $\pinnweights$ (typically using gradient descent), one can therefore find a model that both fits the data and satisfies the physics equation. 
However, so far, the training of $\theta$ must be done separately for each drug configuration $\dinput$. 
We now depict a unified framework for all drug configurations.

Note that the choice of the weights $\lw{Data},  \lw{IC}$ and $ \lw{ODE}$ will affect the training. This choice is model-driven and is discussed for our case-study in Section \ref{sec:choosing:weights}.
Moreover, for sake of simplicity, we wrote the ODE loss  with a common weight for all of its 14 components (given by our CAP model), but it is straightforward to have different weights for the different ODEs (which turned out to be necessary in our case-study).

\subsection{HyperPINN}

We now come back to our original objective which is to learn the function $u(t, \dinput)$ given by the CAP model of Section \ref{sec:CAP:model}, for the drug configuration $\dinput$.

Once a good PINN architecture has been chosen a natural approach would be to learn a mapping which, for a configuration $\dinput$, gives the parameters $\pinnweights(\dinput)$ for which $\approxU(t, \pinnweights(X))$ is a good approximation of $u(t, \pinnweights)$.
This approach, named hyperPINNs, was introduced by \cite{de2021hyperpinn}.
It combines physics-informed neural network architectures with  hypernetworks \citep{ha2016hypernetworks}. 
The main idea is then to build a neural network $\approxU[\text{Hyper}](x, \hyperweights)$, parameterized by $\hyperweights$, that takes as input a drug configuration and  outputs a parameter $\pinnweights$ which serves at the weights for a neural network  $\approxU[PINN](t, \cdot)$. 
Formally, we approximate $u(t, \dinput)$ by setting, for every $(t, \dinput) \in [0,T] \times \mathbb{R}^{5}$: 
\begin{align*}
   \pinnweights(\dinput) &:= \approxU[Hyper]\left(\dinput, \hyperweights\right)\,, \\
    \hat{u}(t, \dinput) &:= \approxU[PINN]\left(t, \pinnweights(\dinput)\right)\,.
\end{align*}
The learning objective then becomes to optimize the loss in $\hyperweights$.
Supposed that we have at disposal a set of $q$ drug configurations $\dinput_1, \dots, \dinput_q$, and that for each configuration, we have a specific set of observations and times for the ODE loss (defined in Section \ref{sec:PINN}).
The loss for the hyperPINN then becomes:
$$
\Loss[HyperPINN](\hyperweights) = \sum_{i = 1}^q \Loss[PINN](\approxU[Hyper](\dinput_i, \hyperweights))\,.
$$
where $\Loss[PINN]$ is defined in Equation \eqref{eq:PINN:loss}. 
 As a hypernetwork, a hyperPINN is expected to have fewer parameters (weights) to learn than the corresponding dense neural network.
 This should then lead to a faster training \citep{chauhan2024brief}.
 Because the ODE system represents a biological process, we refer to the hyperPINN as a hyperSBINN in the following. Once trained, the hyperSBINN acts as a quasi-instantaneous solver of the ODE system for any input configuration $\dinput$, effectively serving as a meta-model by replacing the numerical solver with a faster alternative. Figure \ref{fig:hyperSBINN} depicts graphically the framework of hyperSBINNs, highlighting that the architecture relies on two neural networks: the hypernetwork that provides weights and biases for the SBINN.
 A  parallel can be drawn with \citet{przedborski2021systems}, who also pursue meta-modeling; however, they employ a single neural network architecture and omit the inclusion of an ODE loss during training.
 Our meta-modeling approach differs from that of \citet{sbinn} and \citet{Daneker2023}, whose objective is to estimate ODE parameters while simultaneously training the SBINN. In their case, the SBINNs provide solutions only in the neighborhood of the estimated parameters, rather than for any configuration.

\begin{figure}[ht]
    \centering
\resizebox{\textwidth}{!}{%
\begin{tikzpicture}[
    node distance=1cm and 1.5cm,
    every node/.style={draw, font=\small, align=center},
    output/.style={draw=blue, rounded corners},
    sbinn/.style={draw, rounded corners},
    connection/.style={-Stealth, thick},
    loss/.style={draw, rounded corners, fill=yellow!10},
    loss_block/.style={draw=purple, rounded corners},
    sum/.style={font=\large, draw=none},
    gradient/.style={draw=red, dashed, rounded corners},
    dashed_connection/.style={-Stealth, thick, dashed, draw=red},
]

\node (hypernetwork_input) [rectangle, output] {ODE \\ Parameters \\ $\dinput$};
\node (hypernetwork) [rectangle, right=of hypernetwork_input, output] {Hypernetwork \\ $\approxU[hyper]\left(\dinput, \hyperweights\right)$};
\node (hypernetwork_output) [rectangle, below=of hypernetwork,output] {Output \\ $\pinnweights(\dinput) $ 
};

\node (pinn) [rectangle, below=of hypernetwork_output, sbinn] {SBINN \\ $\approxU[SBINN]\left(t, \pinnweights(\dinput)\right)$};
\node (input) [rectangle, left=of pinn, sbinn ] {Input \\ $\mathsf{u_0}, \obs_{1:n}, t_{1:n}$};
\node (solutions) [rectangle, right=of pinn] {Solutions \\ $\hat{u}(t, \dinput)$};

\node (autodiff) [rectangle, right=of solutions, draw=purple, fill=purple!10] {Automatic \\ Differentiation};
\node (loss_ode) [rectangle, right=of autodiff, loss_block] {$\Loss[ODE](\pinnweights)$};
\node (plus1) [below=0.3cm of loss_ode, sum] {$+$};
\node (loss_data) [rectangle, below=of loss_ode, loss_block] {$\Loss[Data](\pinnweights)$};
\node (loss_ic) [rectangle, above=of loss_ode, loss_block] {$\Loss[IC](\pinnweights) $};
\node (plus1) [above=0.3cm of loss_ode, sum] {$+$};
\node (loss_total) [rectangle, above=of loss_ic, loss] {$\Loss[PINN](\pinnweights)$};
\node (equals) [below=0.3cm of loss_total, sum] {$=$};
\draw[yellow, thick] ([xshift=-0.6cm]loss_total.north west) -- ([xshift=-0.6cm]loss_data.south west) -- ([xshift=0.6cm]loss_data.south east) -- ([xshift=0.6cm]loss_total.north east) -- cycle;


\node (gradient_update) [rectangle, right=of hypernetwork, gradient] {Gradient \\ update};
\node (minimize) [rectangle, right=of gradient_update, gradient] {Minimize};


\draw[->, blue] (hypernetwork_input) -- (hypernetwork);
\draw[->, blue] (hypernetwork) -- (hypernetwork_output);
\draw[connection] (input) -- (pinn);
\draw[connection] (pinn) -- (solutions);
\draw[connection, blue, dashed] (hypernetwork_output) -- (pinn);

\draw[connection] (solutions) -- (autodiff);
\draw[connection] (autodiff) -- (loss_ode);
\draw[connection] (solutions) |- (loss_ic);
\draw[connection] (solutions) |- (loss_data);
\draw[dashed_connection] (loss_total) |- (minimize);
\draw[dashed_connection] (minimize) -- (gradient_update);
\draw[dashed_connection] (gradient_update) -- (hypernetwork);

\end{tikzpicture}

    }
    \caption{The two-stage neural network architecture of hyperSBINN. The hypernetwork takes a set of configuration $\dinput$ as input and outputs the weights and biases for the main PINNs network. The PINNs network then takes a time coordinate as input and makes predictions based on the problem defined by the hypernetwork output.}

    \label{fig:hyperSBINN} 
\end{figure}

\section{Experimental settings}
\label{sec:experiment:settings}

We aim at training  an hyperSBINN to predict CAP at various concentrations. 
Our input $\dinput$ is a vector of size $5$, stacking the IC50 values of four cardiac ion channel targets of interest and the drug concentration.
The model then outputs 14 curves, and in particular the curve of $V(t)$, which is our main target.

We assess the performance of our approach by first comparing the curves outputted by the hyperSBINN to the curves outputted by the solver and second, by computing a quantity of interest (biomarker) from the voltage curve, namely APD90. 

\subsection{Assessing performance of the hyperSBINN model}
\label{sec:assess:perf}

To assess the performance of the hyperSBINN model to retrieve the solution of our ODE system, we compute a normalized root mean square error (NRMSE) over a testing set of curves obtained through an ODE solver.
Our testing set consists of 500 solutions for 500 different drug configurations that were not seen during the training.
These configurations include uniformly simulated IC50 values for four ionic channels, ranging from  $0.1 \, \mu M$  to  $100 \, \mu M$ , and drug concentrations  $\conc$ varying between $ 0 \, \mu M $ and $ 4 \, \mu M $.
For each test configuration $\dinput_i,  (1\leq i \leq 500)$, the solver  outputs a function $u^*(t,\dinput) = \left(u^*_j(t, \dinput)\right)_{1\leq j \leq 14}$ at testing times $t$ evenly spaced between $t_0 = 0$ and $t_{5000} = 500$. We denote $\hat{u}_j(t_k, \dinput_i)$ the output of the hyperSBINN model for channel $j$, drug configuration $\dinput_i$ and time $t_k$. The NRMSE is then computed for each test configuration and each channel as follows:
$$
\mathsf{NRMSE}(i, j) = \frac{1}{r_j}\times \sqrt{\frac{1}{5001} \sum_{k = 0}^{5000} \left(\hat{u}_j(t_k, \dinput_i) - u_j^*(t_k, \dinput_i)\right)^2}\,,
$$
where $r_j$ is a normalizing term to allow comparison between channels. 
Formally, $r_j$ is the spread of possible values for the $j$-th channel (1 for all channels but the voltage, whose spread is between -80 and 80). 
In purpose of comparison, we also compute this metric with three alternatives:
\begin{itemize}
    \item The same hyperSBINN architecture trained without the system biology losses. The training only relies on the training observations given by the solver.
    \item The same hyperSBINN architecture trained without the observation loss. The training only relies on the system biology information, \textit{i.e.}, on the ODE and initial condition losses.
    \item An SBINN without the hypernetwork structure similar to \citet{przedborski2021systems}. We train a single neural network whose inputs are given by $(\dinput, t)$ and all the weights are to be learned. In the following, this method is referred to as metaSBINN.
\end{itemize}

\subsection{Learning specific biomarker of the CAP}

As mentioned previously, the curve $V(t)$ is the main target output. From this curve, we can extract significant  biomarker responses such as the APD50, the peak value, and others. 
An important biomarker is  the APD90 response, which is defined as the time it takes for the membrane potential to repolarize to $90\%$ of its peak value during an action potential. Mathematically, it can be expressed as:
\begin{equation*}
    \text{APD90}= t_{90} - t_{\text{peak}},
\end{equation*}
where $t_{90}$ is the time at which the membrane potential $V(t)$ repolarizes to $90\%$ of its peak value, and $t_{\text{peak}}$ is the time at which $V(t)$ reaches its peak value. These values are  obtained directly from the $V(t)$ curves.
To evaluate the effectiveness of the hyperSBINN model to predict this biomarker, we  compare its predictions of APD90 values across a set of the same drug configurations  as in the previous part. 
The hyperSBINN model was used to predict the voltage curves $V(t)$, from which the APD90 values were subsequently extracted.
The absolute difference with the numerical solver (considered as the ground truth) is then computed for each configuration in the test set.
Additionally, we will compare the APD90 predicted by the hyperSBINN with those obtained from a Random Forest model, a standard machine learning technique which has provided the best performances in this case.
While the methodologies differ, the comparison aims to evaluate performance of predicting the APD90. The Random Forest model will be trained using the same training set as the hyperSBINN (see Section \ref{sec:CAP:model}), and  will  predict APD90 values directly, bypassing the voltage curve $V(t)$ and the CAP model.

\section{Implementation details}

In this section, we provide details of our implementation hyperSBINN. 
\texttt{Python 3.11} was used for all numeric computations.  
HyperSBINN model was designed, trained and deployed using the Python library \texttt{jinns v1.6.1}~\citep{gangloff2024jinns} which uses \texttt{JAX} as a backend.
Note that the training of the hyperSBINN is time demanding, its evaluation (once trained) is fast.

\subsection{HyperSBINN architecture and training}

For the SBINN, a Multi-Layer Perceptron (MLP) architecture with one input the time space $t$ , five hidden layers with 50 neurons each, and an output layer of size 14. 
The hypernetwork is also a MLP with input of size 5, and five hidden layers of 46 neurons. 
The output size corresponds to the required number of parameters for the SBINN network.
The choice of these architectures was guided by different experimentations, where we  tested different combinations to find the optimal configuration that balances computational efficiency and model performance. 
The specific architecture was selected based on its ability to capture the complexity of the CAP model while maintaining manageable computational costs.

Regarding the activation function, we use the hyperbolic tangent activation function for all layers except the last one. The output layer matches the possible values taken by the solutions of the ODEs, which is [0, 1] for the 13 gate functions and $]-\infty; +\infty[$ for the voltage function.
The match is made by using sigmoid and linear activation functions respectively.
For the hypernetwork, we consistently use the hyperbolic tangent activation function.

For stochastic gradient descent, we used the Adamax optimizer \citep{kingma2014adam},  a variant of the Adam optimizer, with a learning rate being constant at $10^{-4}$ for $10,000$ iterations,  then decreasing linearly to $10^{-6}$ until iteration 50000, and being constant to this value for remaining iterations.  
This optimizer was chosen due to its suitability for problems with potentially very large gradients, which can be common in deep learning tasks. 
Unlike Adam, which uses an exponentially decaying average of squared gradients, Adamax utilizes the infinity norm (maximum absolute value) of past gradients. 
In our case, this optimizer was essential as it effectively handled gradients that fluctuate significantly, leading to more stable and efficient learning in our model.

\subsection{Collocation points}
\label{sec:coloc:points}

The ODE loss is computed on a set of 200, 000 times sampled uniformly between 0 and 500, and 200, 000 drug configurations ($\dinput$).
For the latter, the sampling is uniform in dimension 5, with the intervals being $[0.1, 100]$ $\mu M$ for the IC50 values, and $[0; 4]$ $\mu M$ for the drug concentration.
This results in $4\times 10^8$ possible collocation points $(t, \dinput)$. 
Each iteration of the stochastic gradient descent is performed on a mini batch of 500 collocation points.

For readers who want to reproduce our study, it is worth noting that  a preprocessing step is necessary to ensure efficient neural network training. We address this by employing a linear scaling mechanism for the time variable $t$. This involves normalization by dividing $t$ by the maximum value of the time domain, $T_{max}$. This transformation $t = t/T_{max}$ compresses the $t$ scale to a more manageable range, typically around O(1), for optimal NN performance. 

\subsection{Synthetic data} \label{sec:observations}

The synthetic data used to compute the observation loss are outputs of a Livermore solver for ordinary differential equations \citep{radhakrishnan1993description}. 
The solution of the ODE is obtained for 500 different IC50 and drug configurations, sampled uniformly in the same space as the test set or the collocation points (Sections \ref{sec:assess:perf} and \ref{sec:coloc:points}).
The solver outputs the solution with time step 0.1 between $t = 0$ and $t = 100$, and with time step 0.5 between $t=100$ and $t= 500$ (the beginning is more finely sampled since the important features of the signal tend to occur during this period).
To avoid overfitting to specific time points, the solver’s solution is then evaluated at 100 times chosen uniformly at random.



\subsection{Losses weights}
\label{sec:choosing:weights}

The weights discussed in Section \ref{sec:PINN} were meticulously selected to ensure balanced contributions from all terms in the loss function. The weights \(\lw{ODE}\) were chosen so that all terms in the ODE loss function contribute at a similar scale, preventing any single ODE from dominating the optimization process and ensuring the model learns from all aspects of this loss function. The detailed method used for selecting these weights is provided in Appendix \ref{Appendix:weights ODE}.

The weight \(\lw{IC}\) for the initial condition loss term was set to 10 to prioritize satisfying the initial points during training, ensuring the model adheres to the starting state of the system. Additionally, the weights for the observational data loss \(\lw{Data}\) were set to 1. This configuration allows the model to learn from the physical dynamics and governing ODEs while still incorporating observations, but without giving them undue influence over the training process.
As described in Section \ref{sec:assess:perf}, we consider two alternatives of the hyperSBINN where, on one hand, the weights $\lw{IC} = \lw{ODE} = 0$, leading to a model learning without biological information (and thus only based on synthetic data), and, on the other hand, $\lw{Data} = 0$, leading to a model learning without synthetic data (and thus only biology informed).

\section{Results}

\subsection{Performance of the hyperSBINN model}

\begin{figure}[htb!]
\centering
      \includegraphics[width=\textwidth]{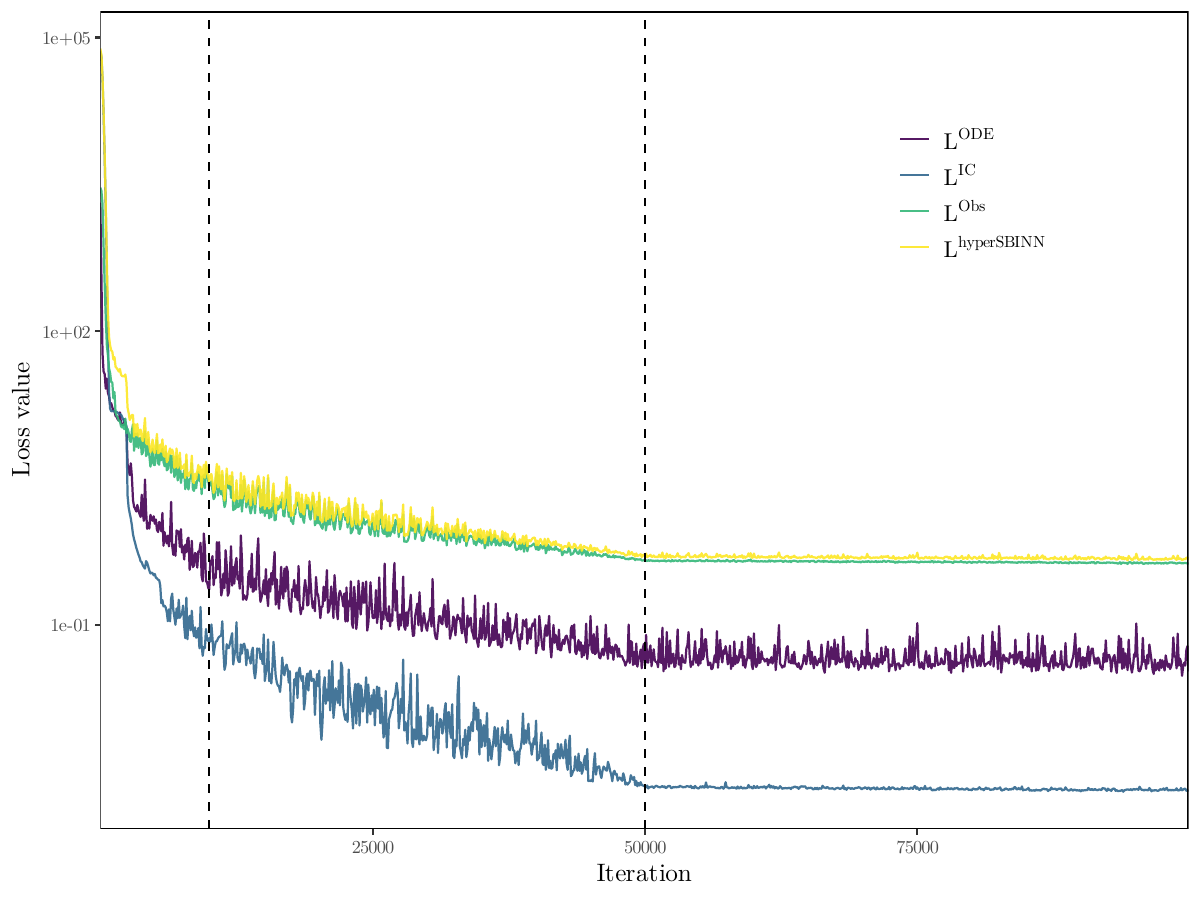}
      \caption{Convergence behavior of different loss components (in log scale) during the training of hyperSBINN. To ease the read, loss values have been smoothed using a moving average over 500 iterations. The learning is constant and equal to $10^{-4}$ until iteration 10000 (first dashed line), then decreases linearly to $10^{-6}$ (second dashed line), and remains constant afterwards.}
    \label{fig:log:loss}
\end{figure}

Figure \ref{fig:log:loss} presents the convergence behavior of the  loss components described in Section \ref{sec:PINN}  during the training of the hyperSBINN model. 
The graph demonstrates the training dynamics of the hyperSBINN, with a significant initial drop in the dynamic loss ($\Loss[ODE]$) indicating rapid learning of the underlying ODEs. 
The steady decrease and eventual stabilization of the initial condition loss and observation loss suggest effective learning of initial conditions and a good fit to the observed data. 
One can notice high variability of the ODE loss, which makes the learning highly sensitive to the learning rate schedule.  
The training of the hyperSBINN takes 18 minutes on a Nvidia T600 laptop GPU with the Python \texttt{jinns} library. 
Then, reconstructing the solution curves, conducted via vectorized pass-forward evaluations in the neural architecture, takes $1.8\times10^{-2}$s. This is approximately $20$ times faster than using the ODE solver ($3.2\times10^{-1}$s for reconstructing the curves).

\begin{figure}[htb!]
\centering
      \includegraphics[width=\textwidth]{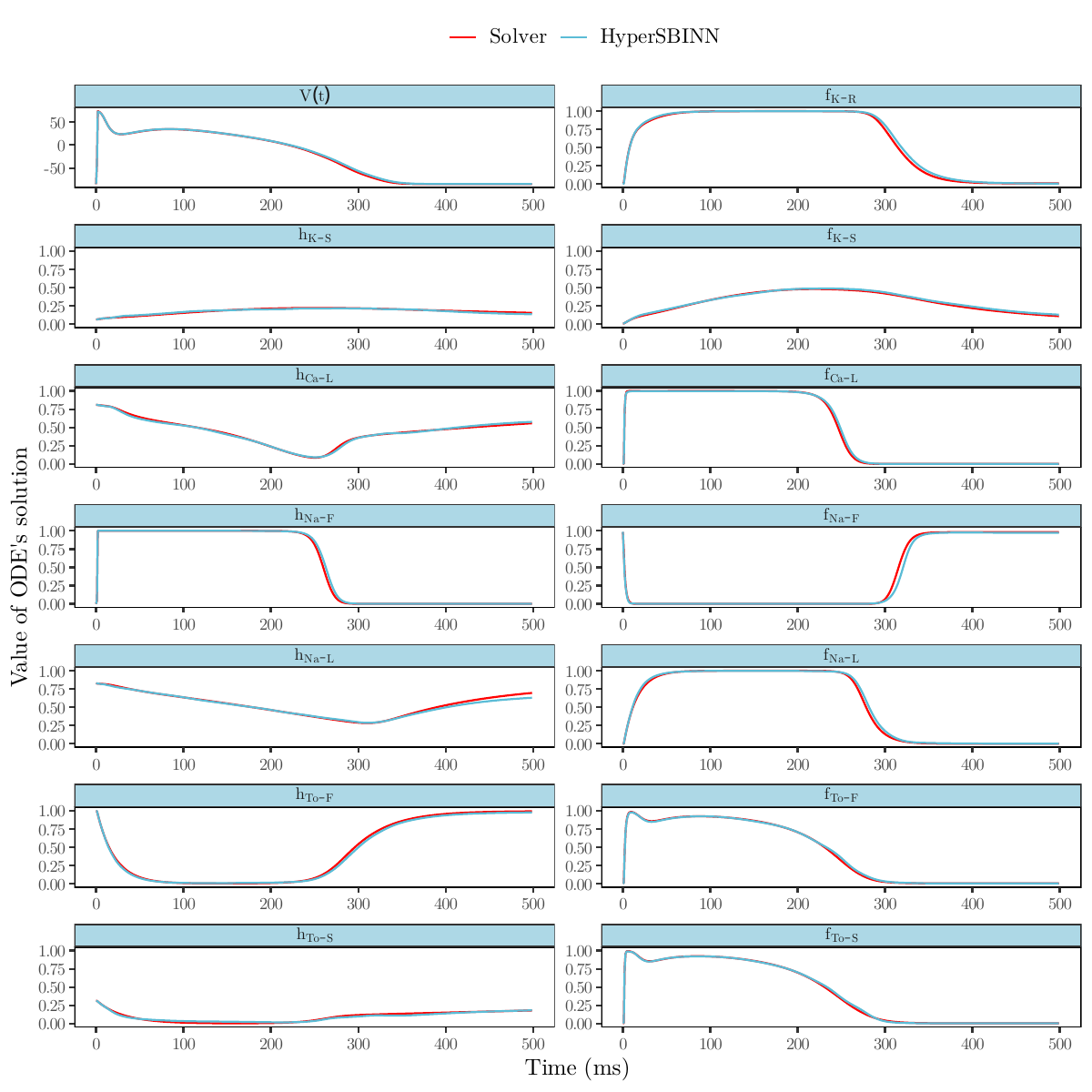}
      \caption{Comparison of the solver and hyperSBINN for a test drug configuration never seen on the training set for each 14 solutions of the biological system. Meaning of all signals names is depicted in Appendix \ref{Appendix:CAP}. Our main signal of interest is $V(t)$ (top left).
      Outputs of our three alternatives to the hyperSBINN are shown in Appendix.
      }
      \label{fig:comparison:solver}
 \end{figure}

\begin{figure}
    \centering
    \includegraphics[width=\linewidth]{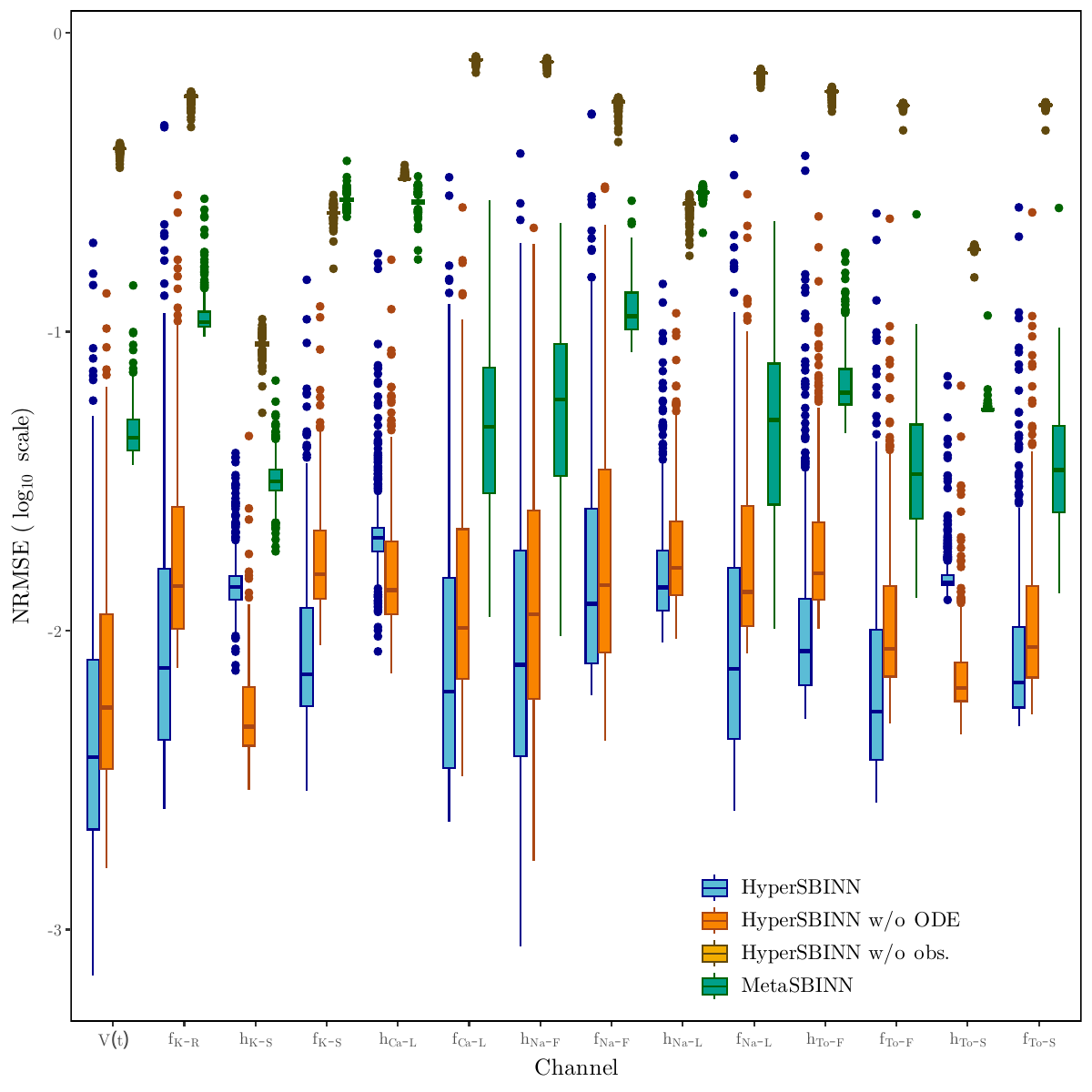}
    \caption{Normalized RMSE (in $\log_{10}$-scale) obtained for each of 14 channels, for 4 different models. Each boxplot is obtained thanks to 500 test curves}
    \label{fig:boxplot:comparison}
\end{figure}

An example output from the hyperSBINN model on a test configuration never seen during the training (specific values are given on Appendix \ref{app:additional:numerics}) on Figure \ref{fig:comparison:solver}. 
One can see a close alignment for the 14 output curves, indicating that the hyperSBINN effectively manages to capture the system's dynamics and can provide comparable performance to the traditional solver. 
The slight discrepancies in signal with low range of variations (such as $h_{\text{Na-L}}$ or $h_{\text{To-S}}$) might be corrected by adjustment of the different ODE weights.
A comparison with the alternatives for the hyperSBINN on this same drug configuration and on another drug configuration is shown on Figures \ref{fig:comparison:all:config1} and \ref{fig:comparison:all:config2}, Appendix \ref{app:additional:numerics}. 
These figures tend to show that both the hyperSBINN with the full loss and the hyperSBINN using only the observations tend to mimick well the ground truth ODE solver, while the two other alternatives provide poor results.
This conjecture is confirmed by Figure \ref{fig:boxplot:comparison} which provides the normalized squared differences between curves outputted by the numerical solver and curves provided by the hyperSBINN on  the test set of 500 drug configurations. 
One can see that for most channels, the hyperSBINN trained with both the biology informed and observation losses is the best. 
However, the hyperSBINN learnt without the ODE loss performs systematically better on three gate functions, namely $h_{\text{K-S}}, h_{\text{Ca-L}}$ and $h_{\text{To-S}}$. 
It is worth noting that these three gate functions are channels whose derivatives tends to be rather small (as shown on an example on Figure \ref{fig:comparison:solver}
Moreover, for three configurations, the hyperSBINN critically fails in comparison to the hyperSBINN without ODE.
This figure also highlights the importance of incorporating ground truth observations in the training, as shown by the poor performances by the hyperSBINN relying only on the biology informed loss.
Finally, one can see that the metaSBINN architecture leads to poorer result than the hyperSBINN architecture. 
This provides a new empirical evidence of the relevance of the hypernetwork.

\subsection{Learning specific biomarker of the CAP}

Table \ref{table:absolute:differences} shows the absolute error of prediction of APD90 when this quantity is computed using the voltage curve output by the hyperSBINN. 
Overall, the prediction is satisfying, as the discrepancy between target and prediction is less than 3.5 ms for 75\% of predictions. 

It is worth noting that if the specific goal is the prediction of this quantity, and if the user has a training set of APD90s for different drug configurations, a classical machine learning algorithm (that focuses on this specific quantity) might be a better option.
For instance, Table \ref{table:absolute:differences} illustrates that a dedicated random forest performs quite well for this task.
This is not a surprise as a perfect prediction of the APD90 from the voltage curves requires a perfect approximation of the lowest value, the highest value, and the decreasing rate. 
A small error on those three quantities could accumulate when predicting the APD90.
Moreover, the hyperSBINN model is prone to extreme values in the error.
Indeed, as one can see on Figure \ref{fig:boxplot:comparison} or on Figure \ref{fig:voltage:curves} (Appendix \ref{app:additional:numerics}), two configurations lead to critically bad prediction of the voltage curve, leading to a catastrophic prediction of the  APD90.
Having a specific sampling design for problematic drug configurations (instead of the uniform sampling chosen here) is certainly a lead for future work.

\begin{table}[ht]
\centering
\small
\begin{tabular}{l|r|r|r|r|r|r}
\hline
 & Min & Q1 & Median & Mean & Q3 & Max\\
\hline
HyperSBINN & 0.0 & 0.600 & 1.400 & 4.140 & 3.400 & 173.600\\
\hline
HyperSBINN w/o ODE & 0.1 & 7.400 & 9.050 & 9.899 & 10.925 & 51.600\\
\hline
HyperSBINN w/o obs. & 299.4 & 310.200 & 311.200 & 313.628 & 313.800 & 394.000\\
\hline
MetaSBINN & 0.1 & 9.675 & 15.850 & 16.306 & 22.600 & 65.700\\
\hline
Random forest & 0.0 & 0.380 & 0.625 & 1.067 & 0.989 & 26.006\\
\hline
\end{tabular}
\caption{\label{table:absolute:differences} Absolute difference (in ms) between predicted and APD90 obtained by the python solver. Predictions are made over 500 drug configurations.}
\end{table}

\section{Conclusion and perspectives}

In this work, we introduce a new framework for modeling cardiac electrophysiology using System Biology-Informed Neural Networks (SBINNs) enhanced with hypernetworks. This  approach addresses the limitations of traditional SBINNs by incorporating hypernetworks for parameterized differential equations, providing a more robust and scalable model of cardiac action potentials. 
Our hyperSBINN approach successfully integrates systems biology-informed neural networks with the flexibility and efficiency of hypernetworks, enabling the simultaneous prediction of cardiac action potentials across drugs at variable concentrations.

This work shows promising results for the metamodelling of a complex biological system.
The hyperSBINN achieves high accuracy in replicating results from traditional solvers for a large range of drug configurations. 
This close alignment underscores the model's capability to effectively capture the dynamics of complex systems.

Although based on underlying biological system,  this work shows that  incorporating synthetic data as observations, even a small portion, significantly improves the accuracy and convergence.  
This is particularly beneficial for complex systems of Ordinary Differential Equations (ODEs) where the model can encounter challenges identifying the correct solution manifold from scratch, as the poor results of the model using only the ODE loss show. 
These observations then act as guideposts, steering the hyperSBINN towards the physically relevant solutions within the vast solution space.

Our results also highlight the relevance of using a hypernetwork, decoupling the role of the drug configuration and time inputs in the predictive model.
In addition to the better performances of the hyperSBINN framework compared to a case were time and drug inputs are equally considered, the hypernetwork modelling has a modular architecture that could ease adaptation of the model to other case study.

However, our hyperSBINN model remains complex, resulting in a training phase that can be long and sensitive to various factors. 
As the system involves numerous interacting equations, and multiple related loss functions, a critical tuning is the choice of loss weights.
A potential improvement would be an adaptive procedure for tuning the loss weights during training, increasing the weights of loss components that have high values or show no decrease. 
This could help the hyperSBINN to have equal performances on all channels, as we showed here that performance over some channels (whose derivatives tend to be small) could be increased.
Such procedure has been proposed by \cite{gao2025physics} for partial differential equations, and would definitely help to have a more stable and easier to tune learning phase.

A second important point is the learning rate schedule which is, as often for such models, critical, as  careful tuning is necessary to balance between exploration and stability during the optimization.

As discussed here, the hyperSBINN framework can be seen as "mesh free" solver. 
In this work, we chose uniform sampling of collocation points both for the solving times of the ODE system, and for the drug configurations.
The results highlight that, while more than often good, the prediction can show bad performances for certain drug configuration.
Uniform sampling, although intuitive, might not be well suited, as it tends to show some aggregating patterns for the sample. 
A potential alternative is to introduce some repulsion using Sobol sampling, which would ensure a larger minimal distance between sampling points, while still offering the exploration performances of random sampling. 
Another attractive method would be an adaptative preferential sampling procedure, which would target configurations (or sampling times) for which the ODE loss shows high values (see \cite{wu2023comprehensive} or \cite{bischof2025multi} and the references therein).
A promising direction for future research would be to carry out a comprehensive comparison between the hyperSBINN framework and operator learning approaches (such as advocated in \cite{huang2024operator}, aimed at assessing the relative benefits and limitations of each method.

To the best of our knowledge, this is the first instance where a system biology-informed neural network is combined with a hypernetwork for the study of cardiac action potentials. 
Looking forward, further refinement and validation of the hyperSBINN model on more specific and diverse datasets are necessary to fully realize its potential. 
Expanding the framework with real clinical and \textit{in vitro} data will be crucial for translating the model advancements into pharmaceutical applications. 
Additionally, improving the selection of observational data could enhance the model performance for practical purposes. 
For instance, incorporating more targeted observations at critical time points or under specific conditions could provide better guidance for the model. 
Furthermore, in our study, we compared our results with the solutions of the ODEs derived from a numerical solver. 
It would be interesting to also compare these results with real cardiac voltage curves and ionic currents derived from biological assays.  
Including real observation data for the gates and voltages would also be beneficial. 
Given that real data often contains noise, methodologies such as those proposed in the work \cite{linka2022bayesian} could be explored. 

Overall, the framework proposed in this work represents a promising step toward more efficient and reliable computational AI models in cardiac electrophysiology, potentially accelerating the development of safer and more effective drugs.

\section*{Acknowledgement}
We thank H. Matter and C. Grebner for thoughtful discussions at the beginning of this project.
 
 \section*{Author Disclosure Statement}
G.H., H.M., M.M., F.S. and J.W. are employees of Sanofi and may hold shares in the company.

 \section*{Funding Information}
	This research was supported by Sanofi.

\section*{Contribution Statement}

\begin{itemize}
    \item I. Soukarieh: Main author, Conceptualization, Methodology, Software, Formal analysis, Writing – original draft, Investigation.
    \item H. Minoux: Supervision, Project administration, Funding acquisition 
    \item M. Mohr, G. Hessler, J. Wenzel: Conceptualization, Methodology, Formal analysis.
    \item F. Schmidt: Conceptualization, Methodology, Formal analysis, Writing-review \& editing.
    \item H. Gangloff: Software, Validation, Writing – review \& editing
    \item P.Barbillon, P. Gloaguen: Conceptualization, Methodology, Formal analysis, Validation, Writing – review \& editing, Supervision
\end{itemize}

\bibliographystyle{plainnat-revised}
\bibliography{biblio}

\newpage
\appendix

\section{Full cardiac action potential model}\label{Appendix:CAP}

This section completes Section \ref{sec:CAP:model}, we give the full equation details for the 8 considered ionic channels in the CAP model.
All constants involved in the following equations are listed on Table \ref{tbl:constant:values}.
In the following, we denote $\sigma(z)$ the sigmoid function
$$
\begin{array}{lccc}
\sigma: & \mathbb{R}  &\rightarrow &  ]0;1[  \\
     & z & \mapsto & \sigma(z) = \frac{1}{1 + \text{e}^{-z}}\,.
\end{array}
$$

The ODE system below depends on five drug characteristics denoted:
$$
\dinput = \left\lbrace \conc, \ICf[\text{K-R}],  \ICf[\text{Ca}], \ICf[\text{To-F}], \ICf[\text{Na-F}]\right\rbrace\,,
$$
where $\conc$ is the concentration of the drug, and the four other terms gives the inhibitory value of the drug on four ionic currents.
To lighten the notation, the dependence on $\dinput$ of the voltage function is omitted, \textit{i.e.} we write $V(t)$ instead of $V(t,\dinput)$.

\subsection{Channels directly impacted by drug configuration}

We first depict the equations for the four currents whose conductance depend on the IC50 value of the tested drug, \textit{i.e.} currents for which the $k_j(\cdot, \cdot)$ function is not equal to 1 in Equation \eqref{eq:genereic:ionic:current}. 

\subsubsection[Fast sodium]{Fast sodium current $I_{\text{Na-F}}$}
$$
    I_{\text{Na-F}}(t, \dinput)= \Gmax[\text{Na-F}] \times  \left(1 + \frac{\conc}{\ICf[\text{Na-F}]}\right)^{-1}\ \times f_{\text{Na-F}}(t) \times h_{\text{Na-F}}(t)\times (V(t)-\Eion[\text{Na-F}])\,,
$$
where
\begin{align}
    \deriv{f_{\text{Na-F}}}(t) &= \frac{\sigma\left(\frac{V(t) + 25}{5}\right) - f_{\text{Na-F}}(t)}{0.005}\,,\label{ODE:f-NaF}\\
    \deriv{h_{\text{Na-F}}}(t) &= \frac{\sigma\left(-\frac{V(t) + 69}{3.96}\right) - h_{\text{Na-F}}(t)}{2}\,.\label{ODE:h-NaF}
\end{align}

\subsubsection[Calcium]{$L$-type calcium current $I_{\text{Ca}}$}

$$
    I_{\text{Ca}}(t, \dinput)= \Gmax[\text{Ca}] \times  \left(1 + \frac{\conc}{\ICf[\text{Ca}]}\right)^{-1}\ \times f_{\text{Ca}}(t) \times h_{\text{Ca}}(t) \times (V(t)-\Eion[\text{Ca}])\,,
$$
where
\begin{align}
    \deriv{f_{\text{Ca}}}(t) &= \frac{\sigma\left(\frac{V(t) + 14.6}{5.5}\right) - f_{\text{Ca}}(t)}{0.7}\,,\label{ODE:f-Ca}\\
    \deriv{h_{\text{Ca}}}(t) &= \left(0.7\text{e}^{-0.0337 \times (V(t) + 14.5)^2} + 0.04\right) \times  \frac{\sigma\left(-\frac{V(t) + 31}{5.54}\right) - h_{\text{Ca}}(t)}{25.1}\label{ODE:h-Ca}\,.
\end{align}

\subsubsection[Fast transient outward]{Fast transient outward potassium current $I_{\text{To-F}}$}
\label{sec:current:TOF}

$$
    I_{\text{To-F}}(t, \dinput)= \Gmax[\text{To-F}] \times  \left(1 + \frac{\conc}{\ICf[\text{To-F}]}\right)^{-1}\ \times f_{\text{To-F}}(t) \times h_{\text{To-F}}(t)\times (V(t)-\Eion[\text{K}])\,,
$$
where 
\begin{align}
    \deriv{f_{\text{To-F}}}(t) &= \frac{\sigma\left(\frac{V(t) + 3}{15}\right) - f_{\text{To-F}}(t)}{3.5\times \text{e}^{-\frac{V^2}{900}} + 1.5}\,,\label{ODE:f-ToF}\\
    \deriv{h_{\text{To-F}}}(t) &= \frac{\sigma\left(-\frac{V(t) + 33.5}{10}\right) - h_{\text{To-F}}(t)}{20\times \left(\sigma\left(-\frac{V(t) + 33.5}{10}\right) + 1\right)}\label{ODE:h-ToF}\,.
\end{align}

\subsubsection[Rectifier potassium]{Rapidly activating delayed rectifier potassium $I_{\text{K-R}}$}
\label{sec:current:KR}

$$ 
I_{\text{K-R}}(t, \dinput) = \Gmax[\text{K-R}] \times  \left(1 + \frac{\conc}{\ICf[\text{K-R}]}\right)^{-1}\ \times f_{\text{K-R}}(t) \times \sigma\left(-\frac{V(t) + 33}{22.4}\right) \times (V(t)-\Eion[\text{K}])\,,
$$
where
\begin{align}
    \deriv{f_{\text{K-R}}}(t) &= 
    \left( 
    \frac{0.00138 \times (V(t) + 7)}{1 - \text{e}^{-0.123 \times (V(t) + 7)}} - \frac{0.00061 \times (V(t) + 10)}{1 + \text{e}^{0.145\times (V(t) + 10)}}
    \right)\label{ODE:f-KR}\\
    & ~~~~~~~ \times \left(\sigma\left(\frac{V(t) + 50}{7.5}\right) - f_{\text{K-R}}(t)\right)\nonumber\,.
\end{align}

\subsection{Channels indirectly impacted by drug configuration}
\subsubsection[Late sodium]{Late sodium current $I_{\text{Na-L}}$}
$$
    I_{\text{Na-L}}(t, \dinput)= \Gmax[\text{Na-L}] \times f_{\text{Na-L}}(t) \times h_{\text{Na-L}}(t)\times (V(t)-\Eion[\text{Na-L}])\,,
$$
where
\begin{align}
    \deriv{f_{\text{Na-L}}}(t) &= \frac{\sigma\left(\frac{V(t) + 30}{5}\right) - f_{\text{Na-L}}(t)}{15}\label{ODE:f-NaL}\,,\\
    \deriv{h_{\text{Na-L}}}(t) &= \frac{\sigma\left(-\frac{V(t) + 75.6}{6.3}\right) - h_{\text{Na-L}}(t)}{120 + \text{e}^{\frac{V(t) + 100}{25}}}\label{ODE:h-NaL}\,.
\end{align}

\subsubsection[Sustained transient outward]{Sustained transient outward potassium current $I_{\text{To-S}}$}
\label{sec:current:TOS}

$$
    I_{\text{To-S}}(t, \dinput)= \Gmax[\text{To-S}] \times f_{\text{To-S}}(t) \times \left(h_{\text{To-S}}(t) + \frac{1}{2}\sigma\left(-\frac{V(t) + 33.5}{10}\right) \right) \times (V(t)-\Eion[\text{K}])\,,
$$
where
\begin{align}
    \deriv{f_{\text{To-S}}}(t) &= \frac{\sigma\left(\frac{V(t) + 3}{15}\right) - f_{\text{To-S}}(t)}{9\times \sigma\left(-\frac{V(t) + 3}{15}\right) + 0.5}\,,\label{ODE:f-ToS}\\
    \deriv{h_{\text{To-S}}}(t) &= \frac{\sigma\left(-\frac{V(t) + 33.5}{10}\right) - h_{\text{To-S}}(t)}{3000\times \sigma\left(-\frac{V(t) + 60}{10}\right) + 30}\label{ODE:h-ToS}\,.
\end{align}

\subsubsection[Slowly activating rectifier current]{Slowly activating delayed rectifier potassium Current $I_{\text{K-S}}$}
\label{sec:current:KL}

$$
    I_{\text{K-S}}(t, \dinput)= \Gmax[\text{K-S}] \times  f_{\text{K-S}}(t) \times h_{\text{K-S}}(t) \times (V(t)-\Eion[\text{K}])\,,
$$
where
\begin{align}
    \deriv{f_{\text{K-S}}}(t) &= 
    \left( 
     \frac{ 7.19 \cdot 10^{-5} \times (V(t) + 30)}{1 - \exp(-0.148 \cdot (V(t) + 30))} +  \frac{1.31 \cdot 10^{-4}  \times (V(t) + 30)}{\exp(0.0687 \cdot (V(t) + 30)) - 1}
    \right)\label{ODE:f-KS}\\
    & ~~~~~~~ \times \left(\sigma\left(\frac{V(t) - 1.5}{16.7}\right) - f_{\text{K-S}}(t)\right)\,,\nonumber\\
    \deriv{h_{\text{K-S}}}(t) &= 
    \frac{1}{4} \times \left( 
     \frac{ 7.19 \cdot 10^{-5} \times (V(t) + 30)}{1 - \exp(-0.148 \cdot (V(t) + 30))} +  \frac{1.31 \cdot 10^{-4}  \times (V(t) + 30)}{\exp(0.0687 \cdot (V(t) + 30)) - 1}
    \right)\label{ODE:h-KS}\\
    & ~~~~~~~ \times \left(\sigma\left(\frac{V(t) - 1.5}{16.7}\right) - h_{\text{K-S}}(t)\right)\nonumber\,.
\end{align}

\subsubsection[Inward rectifier current]{Inward rectifier current $I_{\text{K-I}}$}
\label{sec:current:KI}

$$
    I_{\text{K-I}}(t, \dinput)= \Gmax[\text{K-I}] \times \sigma\left(-\frac{V(t) + 92}{10}\right)\times (V(t)-\Eion[\text{K}])\,.
$$

\begin{table}[ht!]
\centering
\begin{tabular}{|c|c|c|}
\hline
\textbf{Constant} & \textbf{Value} & \textbf{Units} \\
\hline
$\Eion[\text{Ca}]$ & 40 & mV \\
$\Eion[\text{Na}]$ & 74 & mV \\
$\Eion[\text{K}]$ & -85 & mV \\
$\Gmax[\text{Ca}]$ & 0.078 & nS \\
$\Gmax[\text{Na-L}]$ & 0.03 & nS \\
$\Gmax[\text{Na-F}]$ & 16.52 & nS \\
$\Gmax[\text{K-R}]$ & 0.03 & nS \\
$\Gmax[\text{K-S}]$ & 0.1505 & nS \\
$\Gmax[\text{K-I}]$ & 0.29 & nS \\
$\Gmax[\text{To-F}]$ & 0.06 & nS \\
$\Gmax[\text{To-S}]$ & 0.02 & nS \\
\hline
\end{tabular}
\caption{Constant values for the ionic currents models.}
\label{tbl:constant:values}
\end{table}

\section{ODE Weights $\lw{ODE}$ Initialization Method} \label{Appendix:weights ODE}

Several methods exists for initializing weights in neural networks. To effectively balance the contributions of each ODE in the loss function, we implement a specific weighting scheme. 
This method ensures that each ODE contributes proportionately  to the ODE loss, facilitating stable and efficient training.  Below is a step-by-step description of the method used:
\begin{itemize}
    \item \textbf{Initial Weight Setting:} Begin by setting all weights to one. This provides a uniform starting point for the weight adjustment process.
    \item \textbf{Calculate initial ODE loss:} The network is run for a single forward pass, and the initial ODE loss is computed. This is done by calculating the loss after zero epochs.
    \item \textbf{Adjust weights for balance:} Adjust the weights such that the contributions from each ODE to the total loss are of the same order of magnitude.     
\end{itemize}
Our system consists of 14 ODEs, each contributing differently to the ODE loss due to the inherent differences in their dynamics and scales. 
Using the method described above, that aims to address these disparities, the value of $\lw{ODE}$ found vary significantly in a range of 0.01 and 10,000. Without such weighting, certain ODEs with naturally larger magnitudes could dominate the loss, leading to suboptimal training and convergence. 
The chosen weights reflect the relative importance and scale of each ODE in the system. By balancing the contributions of each ODE, the training process becomes more stable and efficient. 
This balance ensures that the optimization algorithm does not focus disproportionately on any single ODE, leading to more uniform updates across the entire system.

\section{Additional numerical experiments}
\label{app:additional:numerics}

\subsection{Example of output for two test configurations}
Here, we provide comparison between the ODE solver and all proposed alternatives for two drug configurations:
\begin{align}
\dinput_1 &=\left\lbrace \conc = 1, \ICf[\text{K-R}] = 12.48,  \ICf[\text{Ca}] = 15.7, \ICf[\text{To-F}] = 9.46, \ICf[\text{Na-F}] = 49.4\right\rbrace\label{eq:config1}\\
\dinput_2 &=\left\lbrace \conc = 0.3, \ICf[\text{K-R}] = 0.48,  \ICf[\text{Ca}] = 75.7, \ICf[\text{To-F}] = 9.46, \ICf[\text{Na-F}] = 49.4\right\rbrace\label{eq:config2}
\end{align}

\begin{figure}
    \centering
    \includegraphics[width=\linewidth]{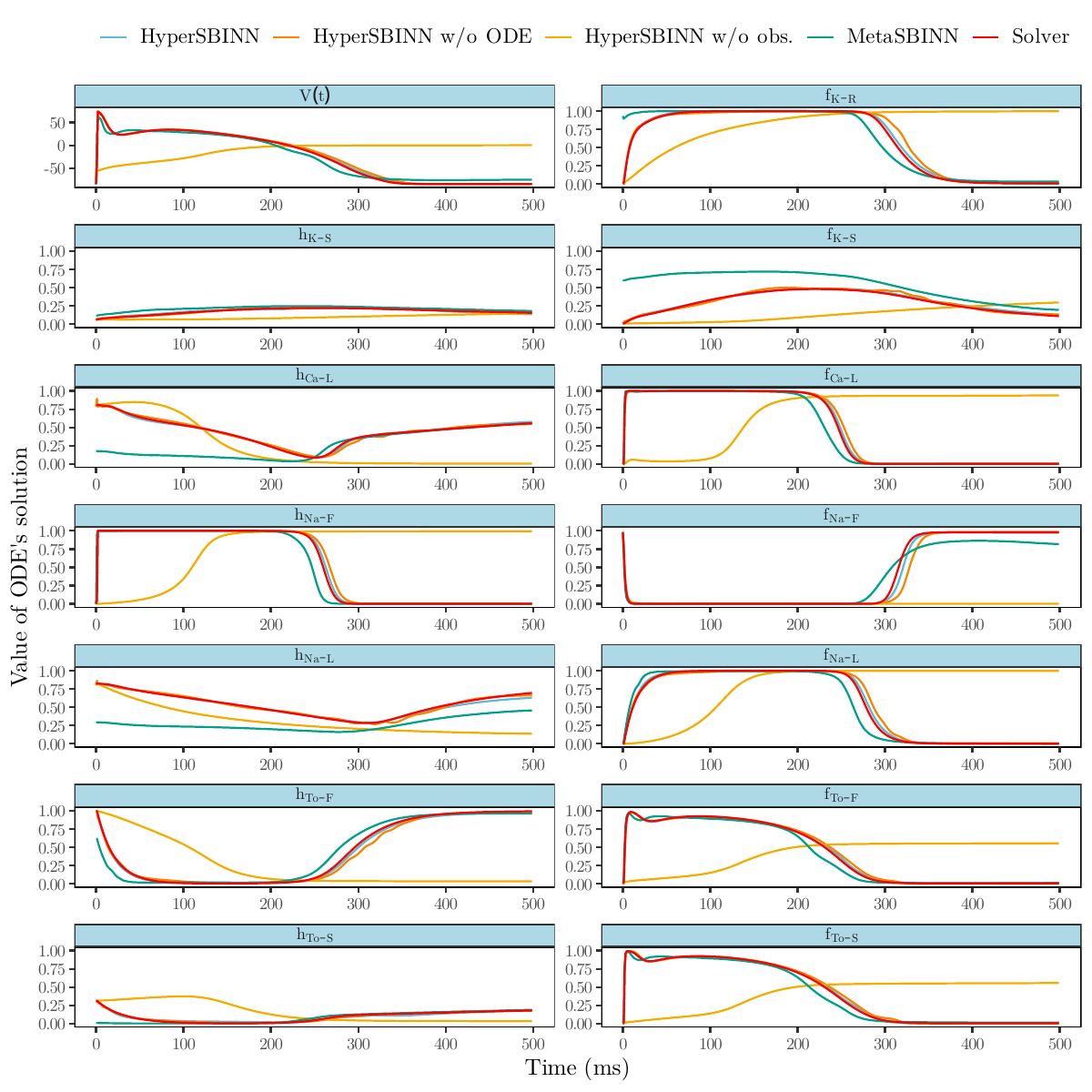}
    \caption{Comparing the performance for drug configuration $\dinput_1$ (Equation \eqref{eq:config1} of the proposed model with with the complete loss (HyperSBINN), without the system biology loss (HyperSBINN w/o ODE), without the addition of actual observations (HyperSBINN w/o obs.), or without an hyper network, \textit{i.e.} with a dense SBINN.}
    \label{fig:comparison:all:config1}
\end{figure}

\begin{figure}
    \centering
    \includegraphics[width=\linewidth]{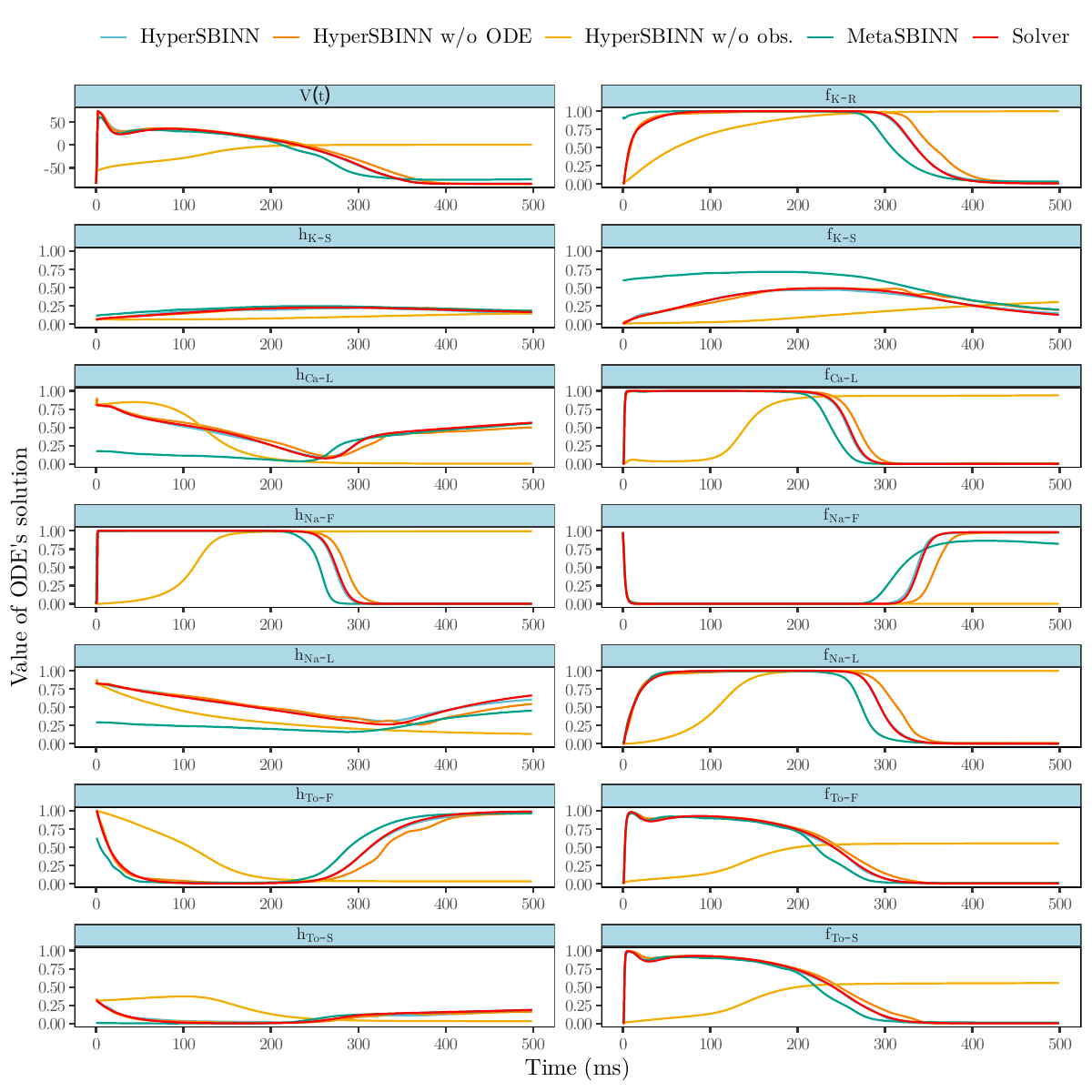}
    \caption{Comparing the performance for drug configuration $\dinput_2$ (Equation \eqref{eq:config2} of the proposed model with with the complete loss (HyperSBINN), without the system biology loss (HyperSBINN w/o ODE), without the addition of actual observations (HyperSBINN w/o obs.), or without an hyper network, \textit{i.e.} with a dense SBINN.}
    \label{fig:comparison:all:config2}
\end{figure}

\begin{figure}
    \centering
    \includegraphics[width=\linewidth]{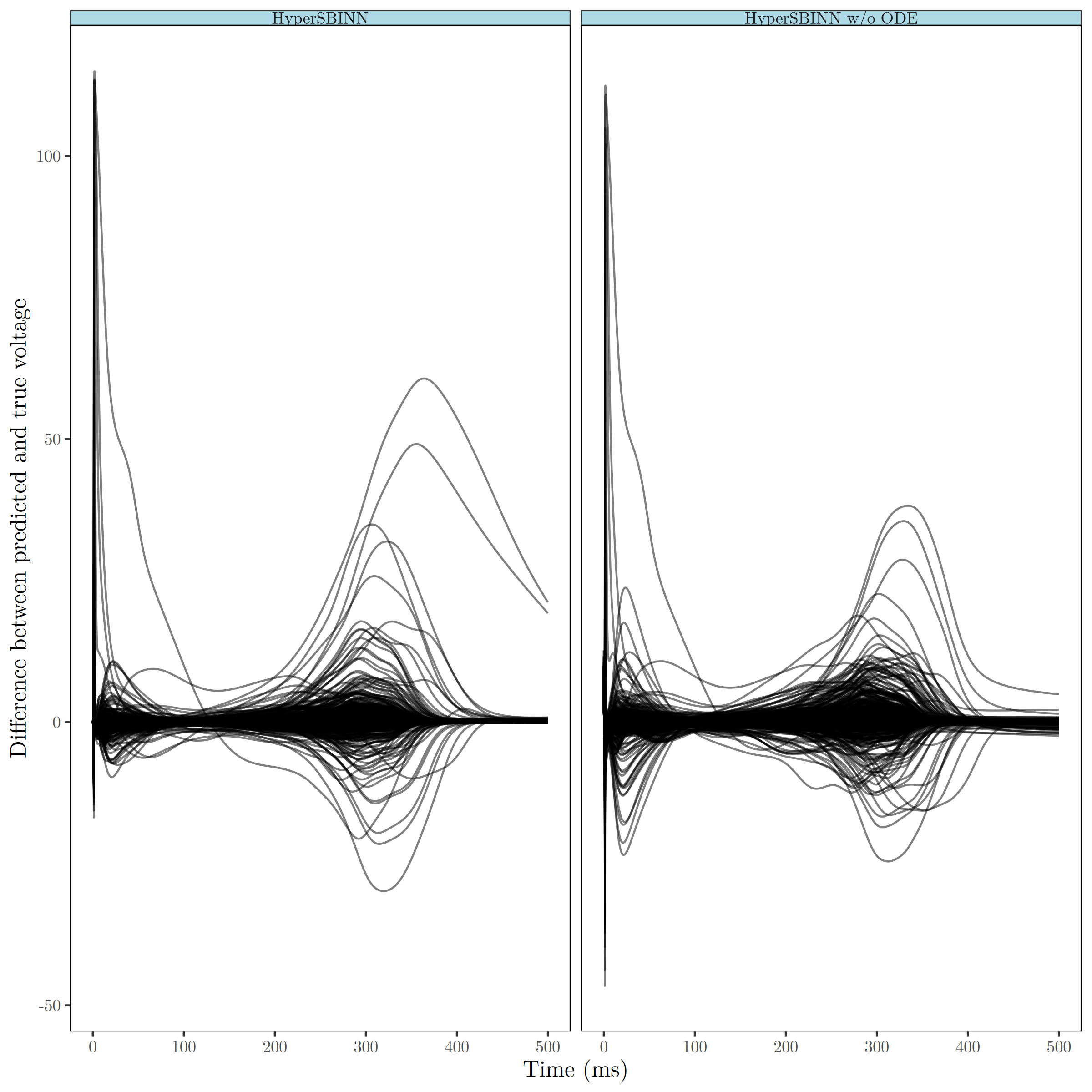}
    \caption{Curves of differences between the predicted and true voltage for the hyperSBINN and its version using only observations. One can see that the hyperSBINN predict two curves which performs really bad after $t = 400$.}
    \label{fig:voltage:curves}
\end{figure}

\end{document}